\documentclass{article}

    \PassOptionsToPackage{numbers, compress}{natbib}

\usepackage[final]{neurips_2021}

\makeatletter
\renewcommand{\@noticestring}{%
      Abstract accepted and presented at the Workshop on Theory of Over-parameterized Machine Learning, April 2021.}
\makeatother

\usepackage[utf8]{inputenc} 
\usepackage[T1]{fontenc}    
\usepackage{hyperref}       
\usepackage{url}            
\usepackage{booktabs}       
\usepackage{amsfonts}       
\usepackage{nicefrac}       
\usepackage{microtype}      
\usepackage{xcolor}         

\usepackage{bbm,xspace}
\usepackage{xcolor,bm,amsmath,graphicx,comment,bigints}
\usepackage{algorithm,algorithmic,subcaption}

\newcommand\cvx{$\mathcal{H}^{tr}$\xspace}

\makeatletter
\newcommand{\printfnsymbol}[1]{
  \textsuperscript{\@fnsymbol{#1}}
}

\title{Deep Learning Generalization, Extrapolation, and Over-parameterization}

\author{%
  Roozbeh Yousefzadeh\\
  Yale Center for Medical Informatics and VA Connecticut Healthcare System\\
  \texttt{roozbeh.yousefzadeh@yale.edu}
}

\begin{document}

\maketitle

\section{Introduction and Summary}
\vspace{-.2cm}
We study the generalization of over-parameterized deep networks (for image classification) in relation to the \underline{convex hull of their training sets (denoted by ${\mathcal{H}^{tr}}$)}. Despite their great success, generalization of deep networks is considered a mystery~\cite{arora2019implicit}. These models have orders of magnitude more parameters than their training samples \cite{neyshabur2019towards}, and they can achieve perfect accuracy on their training sets, even when training images are randomly labeled, or the contents of images are replaced with random noise \cite{zhang2016understanding}. The training loss function of these models has infinite number of near zero minimizers, where only a small subset of those minimizers generalize well \cite{neyshabur2017exploring}. 
Overall, it is not clear why models need to be over-parameterized, why we should use a very specific training regime to train them, and why their classifications are so susceptible to imperceivable adversarial perturbations (phenomenon known as adversarial vulnerability) \cite{papernot2016limitations,shafahi2018adversarial,tsipras2018robustness}. Some recent studies have made advances in answering these questions \cite{belkin2019reconciling,liu2020toward}, however, they only consider interpolation. We show that interpolation is not adequate to understand generalization of deep networks and we should broaden our perspective.

A trained network is a classification function \cite{strang2019linear} that partitions its domain and assigns a class to each partition. Partitions are defined by the decision boundaries and so is the deep learning function. During training, decision boundaries are defined in the domain such that they comply with the labels of training data. As such, the location of decision boundaries in \cvx can be explained by the nearby training images. But, we show that all testing images of standard datasets are outside the \cvx and their directions to \cvx relate to important features in images. Hence, the testing accuracy of models depend on how their decision boundaries extend outside their \cvx. But, how do we shape the extensions of decision boundaries of a model outside the \cvx during training? The answer is \textbf{over-parameterization} in tandem with \textbf{training regime}. We need the model to be over-parameterized in order to have control over the extensions of its decision boundaries, and we should use a specific training regime in order to shape those extensions desirably.

This perspective is novel and has not been considered to explain the generalization of over-parameterized deep networks. Since many recent studies have used polynomials and regression models to explain the generalization of deep networks \cite{belkin2018understand,belkin2018overfitting,belkin2019reconciling,liang2020just,verma2019manifold,kileel2019expressive,savarese2019infinite}, we consider a polynomial decision boundary and rigorously show that we need over-parameterization in order to shape its extensions outside the \cvx. We then study the effect of number of parameters on the output of a neural network in under-parameterized and over-parameterized regimes.

\vspace{-.2cm}

\section{Geometry of image classification datasets}
\vspace{-.15cm}
\paragraph{Domain: Pixel Space.}
For MNIST and CIFAR-10 datasets, we see that all testing images are outside their corresponding $\mathcal{H}^{tr}$. We also see that the distribution of distances to \cvx resemble a normal distribution for both datasets. Moreover, we see that the shortest direction that brings each testing image to the \cvx relates to important features in images. For CIFAR-10, diameter of \cvx is 13,621 (measured by Frobenius norm in pixel space). On the other hand, the distance of farthest testing image from the \cvx is about 3,500 (about 27\% of the diameter). Overall, the distance of testing data to \cvx cannot be dismissed as a small noise.

\vspace{-.5cm}
\paragraph{Wavelet transformation of images.}
When we convolve the images with Haar or Daubechies \cite{daubechies1992ten} wavelets (an operation analogous to convolutions performed by CNNs), the testing samples still remain outside the \cvx in the wavelet space. When we choose a small subset of wavelet coefficients for all images, still, the testing samples remain outside the \cvx and their distances still resemble a normal distribution.

\vspace{-.5cm}
\paragraph{Internal representation of images learned by deep learning.}
The success of deep learning models is attributed to the features they extract from images before their final classification layer. We consider a standard ResNet model \cite{he2016deep} trained on the CIFAR-10 training set. The internal layer of network, just before the last pooling layer, has 4,096 dimensions. In that space, testing samples are still outside the convex hull of training set. 
The last pooling layer then projects the images to a 64 dimensional space, and consequently to a 10-dimensional space which corresponds to the 10 output classes of network. In the 64-dimensional space, all testing samples are outside the \cvx while their distances are significant compared to size of \cvx, and they resemble a normal distribution. 

It may be speculated that there is a low dimensional manifold, learned by deep networks, where all (or most) testing samples are included in the convex hull of training set and the functional task of a deep network on that manifold is interpolation. We observed that there is no trace of such manifold even in the internal representations of a trained ResNet.


\vspace{-.5cm}
\paragraph{Random data in high-dimensional space.}

If we consider random points in the same high-dimensional domain of CIFAR-10, (i.e., domain with 3,072 dimensions bounded between 0 and 255), the testing data will be outside the \cvx and much farther away. That is because the relationship between our training and testing sets are not random and for each testing image, there are a few similar training images, much closer than a random point would be. For the non-random data, we investigate which training images form the surface of \cvx closest to each testing image.

\vspace{-.5cm}
\paragraph{Takeaways from geometry of data.} We can conclude that with such number of training samples in such high-dimensional domains, one can expect the testing samples to be outside their \cvx. However, directions to the convex hulls are usually significant in size, and they contain information related to the object of interest in images. Hence, the generalization of over-parameterized image classifiers cannot be explained by mere interpolation. Instead, we have to consider interpolation in tandem with extrapolation. Interpolation defines the decision boundaries inside the \cvx and brings them to the surface of \cvx. Extrapolation helps us shape their extensions outside the \cvx.

It has been argued that location of decision boundaries may not be important in the pixel space, but how can the domain of our function not be important? Previously, \cite{elsayed2018large,marginbased2019} have shown that pushing the decision boundaries away from the training samples, in the pixel space, improves the generalization of models. Adversarial examples are also created in the pixel space. Another line of research is trying to push the decision boundaries away from samples to make the models less vulnerable \cite{cohen2019certified}. Therefore, the training samples, testing samples, and decision boundaries of the models are closely related to each other in the pixel space, and the location of decision boundaries in the domain actually explain the main functional traits of our deep networks.



\vspace{-.2cm}

\section{Learning outside the convex hull of data: A polynomial decision boundary}
\vspace{-.1cm}

We study a polynomial decision boundary to separate 2 classes of points. Using the orthogonal system of Legendre polynomials, we show that we must over-parameterize the polynomial to have control over its extensions outside the \cvx (over-parameterization is necessary). And we show that once the polynomial is over-parameterized, we need to train it with a specific training regime in order to obtain the decision boundary that we want. This broadens the perspective of previous studies that use interpolation and regression models in order to explain deep learning generalization.

\vspace{-.1cm}

\section{Effect of over-parameterization on the output of neural networks}
\vspace{-.1cm}

We see that in the under-parameterized regime, the main struggle is to minimize the training loss and find the best shape of the decision boundaries inside the \cvx. In the over-parameterized regime, decision boundaries perfectly separate the data inside the \cvx, but depending on the training regime, they can have very different shapes outside the \cvx, showing that over-parameterization gives us control over the behavior of model outside \cvx by using specific training regimes.

\bibliographystyle{plain}
\bibliography{main}

\end{document}